\documentclass[letterpaper, 10 pt, conference]{ieeeconf}  

\IEEEoverridecommandlockouts                              

\usepackage{graphics} 
\usepackage{mathptmx} 
\usepackage{times} 
\usepackage{amssymb}  
\usepackage{xcolor}
\usepackage{hyperref}
\usepackage[style=ieee,natbib]{biblatex}
\usepackage{hyperref}
\usepackage{url}

\usepackage[pdftex]{graphicx}
\usepackage{listings}
\usepackage{xcolor}
\usepackage{amsmath}
\usepackage{wrapfig}
\usepackage{fvextra}
\usepackage{verbatimbox}
\usepackage[]{mdframed}
\usepackage{caption}
\usepackage{subcaption}
\usepackage{svg}
\usepackage{parskip}
\usepackage{soul}

\makeatletter
\let\old@lstKV@SwitchCases\lstKV@SwitchCases
\def\lstKV@SwitchCases#1#2#3{}
\makeatother
\usepackage{lstlinebgrd}
\makeatletter
\let\lstKV@SwitchCases\old@lstKV@SwitchCases

\lst@Key{numbers}{none}{%
    \def\lst@PlaceNumber{\lst@linebgrd}%
    \lstKV@SwitchCases{#1}%
    {none:\\%
     left:\def\lst@PlaceNumber{\llap{\normalfont
                \lst@numberstyle{\thelstnumber}\kern\lst@numbersep}\lst@linebgrd}\\%
     right:\def\lst@PlaceNumber{\rlap{\normalfont
                \kern\linewidth \kern\lst@numbersep
                \lst@numberstyle{\thelstnumber}}\lst@linebgrd}%
    }{\PackageError{Listings}{Numbers #1 unknown}\@ehc}}
\makeatother

\definecolor{codegreen}{rgb}{0,0.6,0}
\definecolor{codegray}{rgb}{0.5,0.5,0.5}
\definecolor{codepurple}{rgb}{0.58,0,0.82}
\definecolor{backcolour}{rgb}{1.,1.,1.}
\definecolor{kaylcolour}{rgb}{0.01,0.98,0.86}
\definecolor{ajcolor}{rgb}{0.1, 0.86, 0.1}
\definecolor{mikecolour}{rgb}{0.01,0.86,0.98}
\definecolor{codehighlight}{rgb}{0.21, 0.9, 0.9}
\definecolor{rebuttalcolor}{rgb}{1.0, 0.0, 0.0}

\lstdefinestyle{mystyle}{
    backgroundcolor=\color{backcolour},   
    commentstyle=\color{codegreen},
    keywordstyle=\color{magenta},
    numberstyle=\tiny\color{codegray},
    stringstyle=\color{codepurple},
    basicstyle=\ttfamily\scriptsize,
    breakatwhitespace=true,         
    breaklines=true,
    breakindent=10pt,
    captionpos=b,                    
    keepspaces=true,                 
    numbers=none,                    
    showspaces=false,                
    showstringspaces=false,
    showtabs=false,                
    tabsize=2,
    frame=single,
    language=Python,
    escapeinside={\%*}{*)},
}
\newcommand{\ours}[0]{GenCHiP}

\title{\LARGE \textbf{
GenCHiP: Gen}erating Robot Policy \textbf{C}ode for \\
\textbf{Hi}gh-\textbf{P}recision and Contact-Rich 
Manipulation Tasks}

\author{Kaylee Burns$^{*1,2}$, Ajinkya Jain$^{1}$, Keegan Go$^{1}$, Fei Xia$^{3}$, Michael Stark$^{1}$, Stefan Schaal$^{1}$, Karol Hausman$^{2,3}$
\thanks{*Correspondence to {\tt\small kayl@google.com}}%
\thanks{$^{1}$[Google] Intrinsic $^{2}$Stanford University $^{3}$Google DeepMind}%
}

\begin{document}

\maketitle
\thispagestyle{empty}
\pagestyle{empty}

\begin{abstract}

Large Language Models (LLMs) have been successful at generating robot policy code, but so far these results have been limited to high-level tasks that do not require precise movement.
It is an open question how well such approaches work for tasks that require reasoning over contact forces and working within tight success tolerances.
%
%
We find that, with the right action space, LLMs are capable of successfully generating policies for a variety of contact-rich and high-precision  manipulation  tasks,  even under noisy conditions, such as perceptual errors or grasping inaccuracies.
%
%
Specifically, we reparameterize the action space to include compliance with constraints on the interaction forces and stiffnesses involved in reaching a target pose.
%
We validate this approach on subtasks derived from the Functional Manipulation Benchmark (FMB) and NIST Task Board Benchmarks.
Exposing this action space alongside methods for estimating object poses improves policy generation with an LLM by greater than 3x and 4x when compared to non-compliant action spaces. 
More material is available on our project webpage: \textcolor{blue}{\href{dex-code-gen.github.io/dex-code-gen/}{https://dex-code-gen.github.io/dex-code-gen/}}
\end{abstract}

\section{Introduction}

Many of the open problems in learning-based robotics revolve around the issue of scaling: deep-learning methods require vast datasets that are not readily available for robotics applications.
One workaround for the data scarcity problem is to retrofit deep learning models that have been trained on internet-scale datasets from other modalities for robotics tasks.
Recently, large language models (LLMs) have emerged as a strong candidate for this approach. LLMs are able to successfully generate code, complete numeric sequences, and solve common-sense reasoning tasks~\citep{codeaspolicies2022, mirchandani2023LargeLM,huang2022language, vemprala2023chatgpt}.
Because code is one of the most popular interfaces for specifying robotic planning and control commands, these capabilities hint at enormous potential when applied to robotics~\cite{codeaspolicies2022}.

Past work demonstrates that generating robot policy code from LLMs is successful for high-level tasks such as navigation and open-vocabulary pick-and-place~\citep{codeaspolicies2022, mirchandani2023LargeLM}. For example, a language model can compose high-level action primitives like {\footnotesize\texttt{grab(chips)}} and {\footnotesize\texttt{move\_to(human)}} to successfully generate a policy conditioned on a natural language command such as ``bring me the chips''~\citep{codeaspolicies2022}.
But at present, lower-level tasks and behaviors are generally considered out of reach for LLMs because, to the best of our knowledge, there are no compelling demonstrations of code generation that commands robots to perform contact-rich tasks. 

\begin{figure*}
    \centering
    \begin{subfigure}{0.6\textwidth}
    \includegraphics[width=\textwidth]{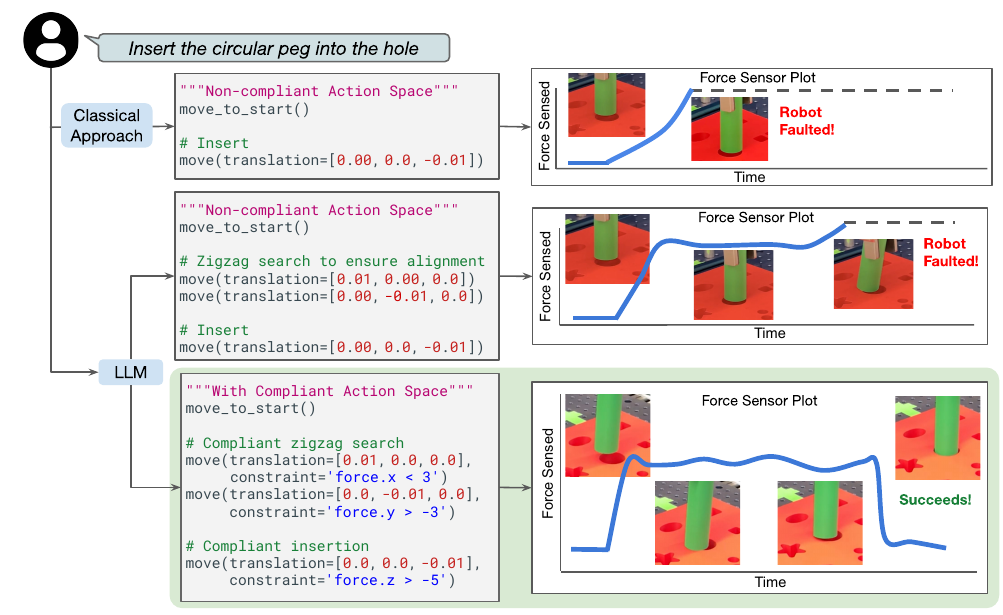}
    \caption{}
    \end{subfigure}
    \hfill
    \begin{subfigure}{0.38\textwidth}
        \includegraphics[width=0.8\textwidth]{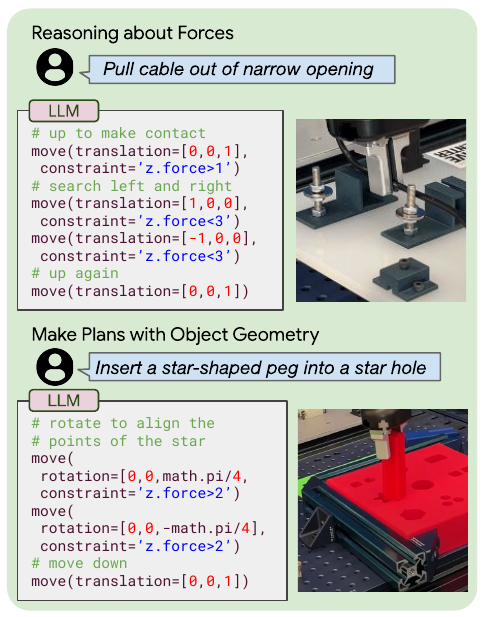}
    \caption{}
    \end{subfigure}
    \caption{\small (a) We prompt an LLM to generate code for high-precision tasks. By using an action space that paramterizes compliant behavior, the LLM is able to generate action sequences for contact-rich tasks like peg insertion. (b) Language models' ability to reason about object geometry and make plans by using world knowledge about different object enables zero-shot generalization to new tasks.}
    \label{fig:intro_figure}
\end{figure*}

While various robot learning approaches have been able to demonstrate impressive generalization across different settings and target objects for pick-and-place tasks~\citep{jiang2022vima, shridhar2022cliport, brohan2022rt, brohan2023rt, shridhar2023perceiver, belkhale2024rth}, such generalization is arguably more difficult for dexterous tasks where a higher level of precision is required.
For example, for a peg-in-hole insertion task, surfaces with more friction or tight insertion tolerances may require multiple insertion attempts or contact force tuning to reach the insertion site~\cite{tracy2023efficient}. Similarly, pegs with different geometries may need different approach trajectories to achieve proper alignment: a peg with a star-shaped cross-section may require an initial rotation for insertion, whereas no such rotation is required for a circular peg.
In practice, the parameters of contact-rich insertion skills are mostly tuned by experts to handle these differences and automating this process is still an open problem.
This presents a challenge for the approach taken in past work~\citep{codeaspolicies2022}, which directly provides a language model with a library of high-level skills (such as {\footnotesize\texttt{insert(peg)}}). 

We present \ours{}: a promising alternative for automating the parameter-tuning process within the control API by instead leveraging the world knowledge inside language models (LMs) to compose lower-level control primitives.
Our goal is to understand if LLMs have the ability to reason about motions and forces acting on objects so as to enable generalization across a larger class of target objects and skills.
To study this topic, we modify the action space in which an LLM operates by exposing constraints on the contact stiffness and forces observed during a high-precision task.
With these modifications we can study code generation for contact-rich manipulation tasks, including industry relevant tasks such as high-precision insertion, rigid body assembly, and deformable object manipulation (see Fig.~\ref{fig:intro_figure}).

The main contribution of this work is to demonstrate that LLMs, without any specialized training, have the ability to perform contact-rich tasks when given the appropriate action space.
We develop a system for automatically generating robot policy code for dexterous tasks by allowing LLMs to specify constraints on the stiffnesses, forces, and trajectories required to perform contact-rich manipulation tasks. We show that our method, \ours{}, is able to outperform a contact-unaware model by over 3x on average on subtasks developed from two challenging contact-rich benchmarks.
Specifically, our approach is able to generate novel insertion patterns from high level descriptions of object shape and texture on insertion tasks from the Functional Manipulation Benchmark (FMB)~\citep{luo2023fmb}, route and un-route cables in the style of the IROS 2020 Robotic Grasping and Manipulation Competition (IROS RGMC)~\citep{Sun2021ResearchCA}, and complete a waterproof connector insertion task selected from the NIST Assembly Task Board \#1~\citep{Kimble2020BenchmarkingPF}.
%
To the best of our knowledge, this is the first paper to provide a strong proof-of-concept that LLMs are capable of parameterizing robotic policies for precise, contact-rich manipulation tasks with code.


\section{Related Work}
\textbf{LLMs for robotics} can successfully generate robot policy code for pick-and-place style manipulation tasks~\citep{huang2022inner, lin2023text2motion}, compose mid-level plans for navigation tasks~\citep{huang2022language}, and compose multiple navigation and manipulation skills for integrated household agents~\citep{singh2023progprompt,wu2023tidybot, xie2023language}. Many of these approaches rely on filtering LLM-generated code based on what is executable~\citep{Gai2021GroundedGD} or on making queries hierarchically~\citep{codeaspolicies2022,belkhale2024rth}. 
They also require a large number of prompt examples: for example, \citet{huang2023voxposer} makes use of eight different prompts with approximately nine subcommands per prompt. By contrast, in the Few-Shot example settings we describe in Sec.~\ref{sec:method-prompting}, we use a maximum of four examples in a given environment.
Furthermore, none of these works establish whether or not LLMs can generate robot policy code for performing high-precision, contact-rich manipulation tasks, which we study in this work.


\textbf{Contact-rich robot manipulation}
tasks are those that involve a robot making controlled contact with its environment while performing them. 
These tasks constitute the vast majority of manipulation tasks in daily life, including household tasks such as wiping tables and sweeping dust into a dust-pan~\citep{wi2023calamari}, and industrial tasks such as high precision insertion~\citep{luo2019icra, zhao2022icra, tracy2023efficient} and assembly~\citep{narang2022factory, liu2022robot}.
A robot needs to reason about the contact forces it will impart on and sense from the environment while performing such tasks to complete them successfully. 
Learning a general policy to perform a wide array of contact-rich manipulation tasks has been studied in great detail in robotics~\citep{kroemer2021review,suomalainen2022survey, elguea2023review, zhao2022icra, morgan2021vision, davchev2022icra, luo2019icra, migimatsu2022icra}, yet how to find a general approach to these tasks remains an open question.
Prior work directly learns policies with imitation learning~\cite{chi2023diffusionpolicy} or reinforcement learning~\cite{Schoettler2019DeepRL,narang2022factory,Brahmbhatt_2023_ICRA}, but these require hundreds of human demonstrations, significant operator training, dedicated simulators, or thousands of environment interactions to achieve a performant policy.
We step towards obtaining a general policy for contact-rich manipulation tasks by leveraging the world knowledge inside LLMs and combining it with the appropriate task action spaces.
We choose robot impedance (or equivalently, admittance) control as the action space for contact-rich manipulation tasks as it can regulate the relationship between robot position and contact forces effectively~\citep{beltran2020variable, abu2020variable}.
\section{Preliminaries}
\label{sec:method-problem}
%
\begin{figure}
\begin{minipage}[t]{\linewidth} 
\centering
\begin{lstlisting}[rulecolor=\color{black},language=Python, linebackgroundcolor={%
    \ifnum\value{lstnumber}=2
        \color{codehighlight!25}
    \fi
    }
    ]
# You're a robot trying to insert a peg in a hole. Grab the circular peg.
pick_up(circular_peg)
\end{lstlisting}
\end{minipage}%
\caption{\small We generate code by formatting natural language requests and instructions as comments. Generations are highlighted in \textcolor{codehighlight}{blue}.}
\label{fig:codeblock-baseline}
\end{figure}
%
Our goal is to develop a system that can translate natural language instructions into robotic actions by leveraging a sufficiently expressive API for control.
Past work~\citep{codeaspolicies2022} shows that off-the-shelf language models can be adapted towards this goal with few-shot prompting. Concretely, pairs of natural language requests with corresponding robot policy code are fed into a language model. Then, the language model can output novel programs in response to new commands as shown in Fig.~\ref{fig:codeblock-baseline}.
The success of this approach can be attributed to the fact that during \mbox{offline} training on vast internet datasets, language models absorb world knowledge about common-sense interactions and learn mappings between natural language instructions and code.
Strategies for adapting this approach towards a contact-rich setting are discussed in the next section.
%

\section{\ours}
\ours{} equips LLMs with a compliant action space. In this section, we first describe the prompting strategies that enable a language model to successfully paramterize and compose compliant move actions. Then, we discuss the different choices of action spaces that can be made available to a language model including our proposed action space. 
%

\subsection{Prompting for contact-rich control}
\label{sec:method-prompting}
We consider four prompting strategies when generating robot policy code from a language model:

a. \textbf{Task descriptions} are high-level explanations of the scene and the task goal written in natural language.
%
\begin{figure}
\begin{minipage}[t]{\linewidth} 
\begin{lstlisting}[title={(a) Task Description}, rulecolor=\color{black},language=Python, linebackgroundcolor={%
    \ifnum\value{lstnumber}>1
        \color{codehighlight!25}
    \fi
    }
    ]
"""You're a robot trying to undo cable routing. Unroute the cable from the screws and brackets it is wrapped around."""
\end{lstlisting}
\end{minipage}%

\begin{minipage}[t]{\linewidth} 
\begin{lstlisting}[title={(b) Control API Descriptions}, language=Python, linebackgroundcolor={%
    \ifnum\value{lstnumber}>100
        \color{codehighlight!25}
    \fi
    }
    ]
"""Use these methods:
- move: moves to specified offset 
  Args:
      translation: (x, y, z) tuple
      rotation: (x, y, z) tuple
..."""
\end{lstlisting}
\end{minipage}%

\begin{minipage}[t]{\linewidth} 
\begin{lstlisting}[title={(c) Hints}, rulecolor=\color{black},language=Python, linebackgroundcolor={%
    \ifnum\value{lstnumber}>100
        \color{codehighlight!25}
    \fi
    }
    ]
"""Rules:
- Don't define any new methods
- Don't call any undefined methods
- Don't add any if statements or while loops"""
\end{lstlisting}
\end{minipage}%

\begin{minipage}[t]{\linewidth} 
\begin{lstlisting}[title={(d) Examples}, language=Python, linebackgroundcolor={%
    \ifnum\value{lstnumber}>100
        \color{codehighlight!25}
    \fi
    }
    ]
# Move the cable to the left until it snags
move((1, 0, 0), 
     constraint=(x.force>-1))
\end{lstlisting}
\end{minipage}%
\caption{\small We present information about the task and control API via prompting. The API description is the same across all environments, the hints and examples are the same within each environment, while the task description must be modified to describe each task.}
\label{fig:codeblock-prompts}
\end{figure}
%
These can occur at both the beginning and end of a prompt and often include important information about the task setup such as the peg shape or the available objects. See Fig.~\ref{fig:codeblock-prompts}a for an example.

b. \textbf{Descriptions of available control APIs} are formatted doc-strings that describe code accessible to the LLM. These include lists of variables as well as the expected range of values for floating point numbers. Fig.~\ref{fig:codeblock-prompts}b shows an example description for a simplified {\footnotesize{\texttt{move}}} function that controls end-effector pose. We also include descriptions for the full library of available methods, including a point-to-point move, a compliant move, conditions, gripper movements, and methods or variables specifying the positions of relevant objects.
% ...

c. \textbf{Hints} in our setting include rules, keywords that specify relevant control primitives, and requests to have the model explain its reasoning in natural language or in pseudocode. Phrases such as ``perform a pattern search'' guide the model towards predicting behavior that better recovers from errors and better handles imprecision in the position of target poses. Intuitively these keywords help in reducing task ambiguity (e.g., by emphasizing that provided locations are imprecise) and guide the model towards motion patterns that are relevant to contact-rich tasks. Requests to explain in natural language can be thought of as a variant of chain-of-thought prompting~\cite{Kojima2022LargeLM}. The specific keywords and requests that are helpful in each task are described in the experimental section.


d. \textbf{Examples} of the control APIs being used for basic movements, such as making contact with a surface, are useful for tasks with ambiguity or where the desired force constraints are difficult to infer from the given ranges.

The combination of the prompt strategies described above allows us to prompt a language model with enough contextual information about the dexterous task at hand. Next, we discuss how we can design the action space of the robot to be able to perform such tasks in practice.

\subsection{Action spaces for robot manipulation tasks}
\label{sec:method-action-space}
Past approaches assume access to a library of methods that exhaustively cover all user-requested commands~\citep{codeaspolicies2022, mirchandani2023LargeLM}.
Building such a library is challenging for contact-rich tasks because in practice these policies are tuned by experts across different object geometries, frictions, and scene layouts.
This section describes different approaches to parameterizing the control API and what the right choice of a control API can achieve.
Formally, we consider a contact-rich robot manipulation task to be composed of a sequence of subtasks $\tau = (t_1, \ldots, t_n)$. The specific definition of a subtask will change based on the action space, as described below. 

\begin{figure}
\begin{minipage}[t]{\linewidth} 
\centering
\begin{lstlisting}[language=Python, linebackgroundcolor={%
    \ifnum\value{lstnumber}>1
        \color{codehighlight!25}
    \fi
    }
    ]
# Insert a peg into a hole
pick_up(peg)
# go down to make contact 
move_point_to_point([0, 0, -1])
# wiggle to find opening
move_point_to_point([1, 0, 0])
move_point_to_point([-1, 0, 0])
# go down to insert
move_point_to_point([0, 0, -1])
\end{lstlisting}
\end{minipage}%
\caption{\small Generating code with point-to-point moves limits policies to free-space-motions. For this policy to run successfully, displacement along the z-axis in the second and fifth actions must be exact.}
\label{fig:codeblock-intro}
\end{figure}

\textbf{Point-to-point moves.} 
Past work~\citep{codeaspolicies2022,mirchandani2023LargeLM} makes use of an action space that directly commands the robot to move to target poses in the Cartesian space, $[\mathbf{x}_{target}]_i$ (See Fig.~\ref{fig:codeblock-intro}).
In this setting, each sub-task $t_i$ is simply defined as the next Cartesian pose (a.k.a. waypoint):
$t_i = ([\mathbf{x}_{target}]_i)$.

While this approach is successful for executing motions in free-space or for simple pick-and-place tasks, it fails when the robot needs to explicitly make a purposeful contact with its environment. Consider a robot trying to make contact with a surface to perform a wiping motion. Successfully parameterizing a policy in this action space would require predicting a precise Cartesian pose with very little tolerance for error. Predicting a pose that is millimeters short of the surface would fail to make a contact and predicting millimeters too deep into the surface would cause the robot to exert high forces on the surface, which can cause faults in the robot or even break it in the worst case scenario.

\textbf{Compliant moves.}
Addressing this shortcoming, we propose to parameterize the action space for performing contact-rich manipulation tasks using a robot's compliance, realized in impedance control (or equivalently admittance control which is an adequate action space for robot learning in~\citet{MartnMartn2019VariableIC}).
An impedance move action is parameterized by both a target Cartesian pose, $[\mathbf{x}_{target}]_i$, and a vector that specifies stiffness along each degree of freedom, $\sigma_i$:
$t_i = ([\mathbf{x}_{target}]_i, \sigma_i )$,
when the robot is in contact with the environment. During execution, the stiffness vector for each subtask can be used to define the parameters for a variable impedance controller~\citep{buchli2011learning} of the form:
\begin{align}
F_{external} = K_p(\mathbf{x}_{target} - \mathbf{x}) + K_d(\dot{\mathbf{x}}_{target} - \dot{\mathbf{x}}) + \Lambda (\ddot{\mathbf{x}}_{target} - \ddot{\mathbf{x}})
\label{eqn:impedance-control}
\end{align}
where $\mathbf{x}_{target}$, $\mathbf{x}$, $\dot{\mathbf{x}}_{target}$, $\dot{\mathbf{x}}$, $\ddot{\mathbf{x}}_{target}$, and $\ddot{\mathbf{x}}$ denote the target and current pose, twist, and accelerations, respectively. $K_p$, $K_d$, and $\Lambda$ correspond to the stiffness, damping, and task-space inertia matrices, respectively. The impedance controller realizes that the end-effector in contact with the environment behaves like the linear spring-damper-mass system above. $K_p$, $K_d$, and $\Lambda$ are computed as a function of our specified stiffness vector $\sigma_i$ (explained below) and robot specific parameters in order to achieve stable yet responsive behavior. 
%
\begin{figure}
\begin{minipage}[t]{\linewidth} 
\centering
\begin{lstlisting}[language=Python, linebackgroundcolor={%`
    \ifnum\value{lstnumber}>1
        \color{codehighlight!25}
    \fi
    }
    ]
# Insert a peg into a hole
pick_up(peg)
# go down to make contact 
move_conditional_compliant([0, 0, -1], 
     constraint=(z.force>1))
# wiggle to find opening; stop when force lessens
move_conditional_compliant([1, 0, 0], 
    constraint=(z.force<1))
move_conditional_compliant([-1, 0, 0], 
    constraint=(z.force<1))
# go down to insert
move_conditional_compliant([0, 0, -1], 
    constraint=(z.force>2))
\end{lstlisting}
\end{minipage}%
\caption{\small Compliance prevents the robot from faulting when in contact. Conditional termination constraints enable the language model to reason about contact forces. In this example, the robot moves a cable back and forth until no more upward force is detected, which indicates that an opening has been found.}
\label{fig:codeblock-impedance}
\end{figure}

%

Intuitively, the stiffness vector determines the interaction forces that the robot will impart on its environment while performing the task. Low stiffness coefficients in $\sigma$ regulate the robot's compromise between contact forces and the attempt to achieve position accuracy. In the example that we discussed in the last paragraph, a low stiffness value would enable the robot to maintain gentle contact with a surface that prevents the robot from reaching a desired position. A higher stiffness value would create higher contact forces, equivalent to a higher priority to reduce position error.

\textbf{Conditional compliant moves.} In addition to the impedance control specification described above, we also allow the LLM to specify conditions under which to terminate an impedance move. Specifically, these are thresholds on force or position in a specified coordinate direction. Example pseudocode is presented in Fig.~\ref{fig:codeblock-impedance}. This is a powerful primitive as it enables the robot to construct recipes for high-precision tasks without relying on fine-grained perception. In the example of making contact with a surface, this may look like moving a peg downwards with a termination constraint on upward force. 

\section{Experiments}
In this section, we evaluate the ability of \ours{} to generate code for fine-grained manipulation tasks that require high precision in a series of experiments on a set of real robotic tasks. Specifically, we evaluate on a subset of high-precision contact-rich manipulation tasks from the Functional Manipulation Benchmark (FMB)~\citep{luo2023fmb} and a set of industrial manipulation tasks selected from the NIST Assembly Task Board \#1~\citep{Kimble2020BenchmarkingPF} and the IROS 2020 Robotic Grasping and Manipulation Competition (IROS RGMC)~\citep{Sun2021ResearchCA} (Sec.~\ref{sec:robot-experiments}). 
Later, we ablate prompt hints to study the utility of incorporating additional hints in generating relevant motion patterns for robot manipulation tasks (Sec.~\ref{sec:hint_ablation}).

\subsection{Contact-rich manipulation tasks}
\label{sec:robot-experiments}
Sec.~\ref{sec:tasks} details the task setup considered in our experiments, while baselines and ablations are discussed in Sec.~\ref{sec:baselines}. We discuss experimental results in Sec.~\ref{sec:results}. \\

\subsubsection{Task description}
\label{sec:tasks}
The \textbf{Functional Manipulation Benchmark}~\citep{luo2023fmb} studies robotic manipulation, grasping, reorienting, and assembling of a set of dozens of 3D printed objects. The benchmark emphasizes generalization across different object shapes and positions. We evaluate our approach on a subset of peg insertion tasks across three different object shapes: the circle, star, and half-pipe. We use scripted motion to bring the pegs into a fixed position over the insertion point and program a randomized rotation about the z-axis. There is no rotation for the circular peg because it has a constant radius. Rotation of the star is sampled uniformly between 0 and $\frac{\pi}{2}$. Rotation of the half pipe is sampled uniformly from either 0 or $\pi$. Inserting these peg shapes successfully requires generating search patterns contingent on object geometry.

\begin{figure*}
    \centering
    \includegraphics[width=\textwidth]{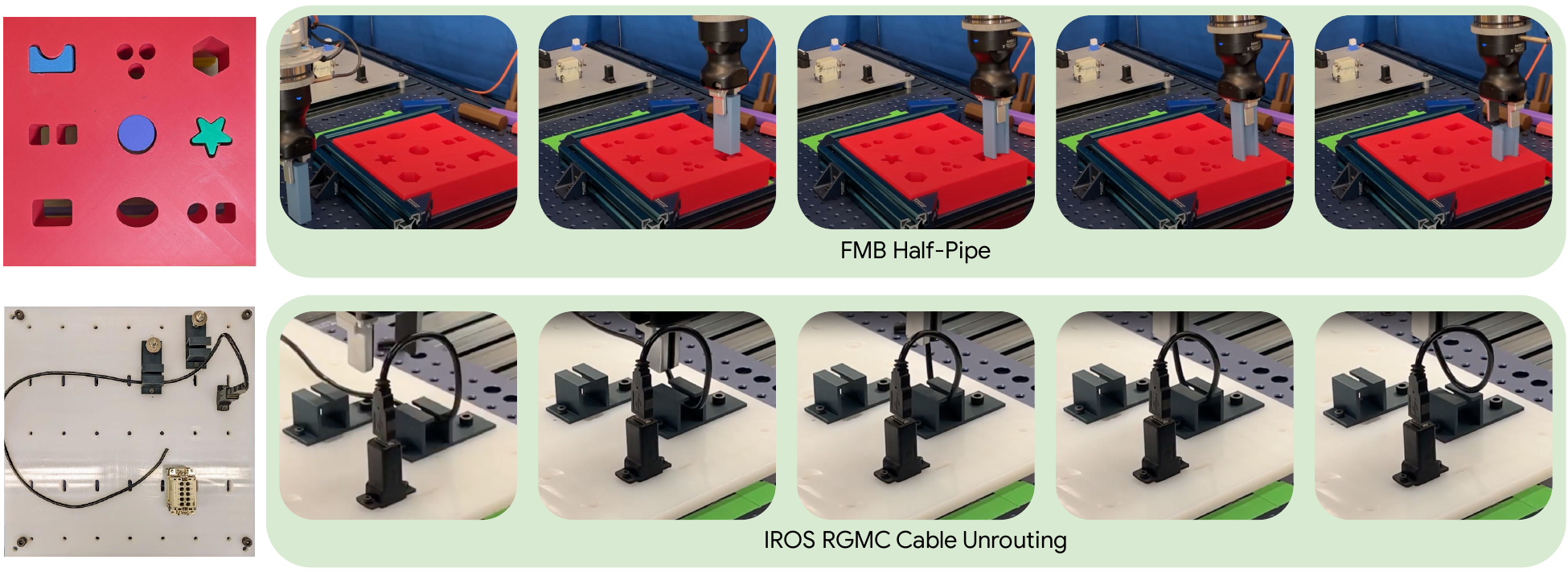}
    \caption{\small Left: The Functional Manipulation Benchmark (FMB)~\citep{luo2023fmb} and a modified assembly board with tasks sourced from NIST Assembly Task Board \#1~\citep{Kimble2020BenchmarkingPF} and IROS RGMCS 2020~\citep{Sun2021ResearchCA} we used for experimentation. Both environments have relatively tight tolerances. Right: Example of the rollouts produced by our method for two tasks: the waypoints from \ours{} are able to successfully complete the tasks.}
    \label{fig:tasks}
\end{figure*}
   
\textbf{Industrial Manipulation Tasks} are adapted from the NIST Assembly Task Board \#1~\citep{Kimble2020BenchmarkingPF} and the IROS RGMC 2020~\citep{Sun2021ResearchCA}.
These benchmarks are designed to evaluate proficiency in robotic assembly with an emphasis on small and medium sized parts and deformable objects.
We consider the wire routing and connector insertion subtasks. Specifically, we study routing (insertion) and unrouting (removal) of a wire through a plastic channel component and inserting a waterproof connector plug into a socket. 
Task environments are visualized in Fig.~\ref{fig:tasks}.
Note there are multiple sources of noise across episodes. For example, in the orientation of the cable within the grasp of the gripper, the tautness of the cable, and the estimated poses of the waterproof connector. Additionally, we evaluate on each side of the routing component to verify that the pattern works across starting positions.
An ideal method should result in an execution policy that is robust to such sources of noise. \\

\subsubsection{Implementation details}~We conduct our robot experiments on a Universal Robotics UR5e robot, which is a position-controlled robot with ATI Axia80 force-torque sensor at the wrist. To expose the compliant action space to the language model, we prompt it with the doc-string for a Cartesian admittance move with parameters on stiffnesses, impedances, and constraints in reaching a target pose. We add a suffix describing the details of the given task, optionally including certain keywords about relevant motion patterns when the environment setup is ambiguous (e.g., we specify that the peg in the FMB insertion tasks is not aligned, which requires the policy to search for the opening).

We give the LLM access to methods for computing transformations on reference poses (i.e., {\footnotesize\texttt{pose\_multiply}}) and for detecting objects in the environment (i.e., {\footnotesize\texttt{estimate\_and\_update\_pose}}). To support the latter, we train pose estimators to do joint object detection and keypoint prediction using a Faster-RCNN~\citep{renNIPS15fasterrcnn} architecture. \\

\subsubsection{Methods considered}
\label{sec:baselines}
We compare two classes of methods: a scripted baseline policy authored by an expert and different variants of LLM-generated code using the prompting strategies and control APIs outlined in Sec.~\ref{sec:method-prompting} and Sec.~\ref{sec:method-action-space}.
For LLM-generated code, we compare with and without code examples. For zero-shot settings, the prompt includes includes Task Descriptions (Fig.~\ref{fig:codeblock-prompts}a), Control API Descriptions (Fig.~\ref{fig:codeblock-prompts}b), and Hints (Fig.~\ref{fig:codeblock-prompts}c). In few-shot settings, Examples (Fig.~\ref{fig:codeblock-prompts}d) are added to the prompt.
%

\textbf{Scripted [Baseline].} We compare against a scripted pattern search insertion move that is tuned by an expert on a single task setting. This baseline reflects an alternative to our approach where a single skill is added to our control library, but is not able to be tuned by an expert across different task generalizations. On FMB tasks, we adapt a pattern search insertion skill for peg insertion. The scripted move implements fixed get-in-contact, pattern search, and insertion phases, with durations, motion patterns, and force thresholds set by an expert on the circle setting.

\textbf{Code-as-Policies (CaP) \citep{codeaspolicies2022} [Baseline].} We compare with a baseline approach akin to the prior work~\citep{codeaspolicies2022, mirchandani2023LargeLM} that uses the point-to-point action space for performing robot manipulation tasks, i.e. to directly command the robot to move to Cartesian target poses (Fig.~\ref{fig:codeblock-intro}). 

\textbf{\ours{}, Fixed Compliance (FC).} For each task, we compare against a baseline where we do not expose the stiffness and impedance targets or the force constraints to the LLM planner, but instead, use predefined compliance parameters. This ablates the importance of force constraints in completing the task, making the action space similar to prior work~\citep{codeaspolicies2022}, but with compliant motions. Concretely, we provide a modified prompt and access to a wrapper around the Cartesian admittance move that provides fixed stiffness and impedance targets and a fixed translation error constraint.


\textbf{\ours{} (Few-Shot).} We expose force constraints to the language model and add examples of calls to our control API, which includes conditional compliant moves. This is similar to the \ours{}, Fixed Compliance baselines, but all of the force constraints and termination conditions are exposed to the language model (Fig.~\ref{fig:codeblock-impedance}). In these experiments, each ``shot'' is an example subcommand that shows how to call the API. In the IROS RGMCS tasks, we include 3 example subcommands: moving down until contact, moving up until a snag, and moving right until a snag is detected.

\textbf{\ours{} (Zero-Shot).} We follow the same approach of exposing force constraints to the language model as \ours{} (Few-Shot, Fig.~\ref{fig:codeblock-impedance}), but do not include any examples that call the control APIs. This is the most difficult generalization setting because every command is an unseen command.

\subsubsection{Evaluation Protocol}
Similar to \citet{yu2023language}, we take the best completion out of 5 calls to the underlying language model and run 10 evaluations. All of our experiments use the OpenAI ChatCompletions API~\citep{Achiam2023GPT4TR} with the {\footnotesize\texttt{gpt-4-0613}} endpoint and a temperature of 0.0. To make comparisons between different action spaces as fair as possible, we take the most successful code generated from our method and overwrite the control API to implement the relevant action space. 
For each environment, we tune the insertion reference pose that appears in the prompt. Concretely, this is the reference pose used in to make contact in the admittance move. 
This hyperparameter is essential on the Point-to-Point baseline because insertion reference poses that are too deep cause a fault. \\

\begin{table}[t]
\begin{center}
\begin{tabular}{c c c c}
\hline
 & Circle  & Star & Half-Pipe \\ 
\hline
Scripted & \textbf{100\%} & 10\% & 0\% \\
CaP \citep{codeaspolicies2022} (Zero-Shot) & 70\% & 0\% & 0\% \\
\ours{}, FC (Zero-Shot) & \textbf{100\%} & 70\% & 30\% \\
\textbf{\ours{}} (Zero-Shot)  & \textbf{100\%} & \textbf{80\%} & \textbf{50\%} \\
\hline
\end{tabular}
\end{center}
\caption{\small \textbf{Functional Manipulation Benchmark.~\cite{luo2023fmb}} Compared to a scripted policy, \ours{} is better able to generalize an insertion pattern across object geometries, including the challenging half-pipe task. We show average performance across 10 evaluations.}
\label{tab:fmb}
\end{table} 



\subsubsection{Results}
\label{sec:results}
\textbf{Functional Manipulation Benchmark.}
Our first evaluation studies how well different approaches generalize across unseen task settings, in particular, the ability to modify insertion search patterns based on different object geometries. Results on FMB are listed in Tab.~\ref{tab:fmb}. 

We find that \ours{} outperforms the performance of other methods across different peg shapes. The baseline scripted policy is successful on the star only when the points are already in close-enough alignment with the hole and fails on the half-pipe shape, which is the most difficult to align because there is only one valid orientation for a successful insertion.
In contrast, our method is successful on half-pipe 50\% of the time. Upon further analysis, we notice that this is because \ours{} generates a successful policy in only one direction of rotation (i.e., 100\% successful for one rotation and 0\% successful for the other randomized rotation). 

When we inspect the code output from the LLM, we find that it generates intuitive waypoints for the search that correspond to the object specified in the prompt. For example, for the half-pipe, the output waypoints oscillate between $0$ and $\frac{\pi}{2}$ while the star shape policy goes through multiples of $\frac{\pi}{4}$. We also find that the zero-shot prompting setup is sufficient for this application. Out of the box, the language model is able to parameterize the conditional compliant move.

\begin{table}
\setlength{\tabcolsep}{2pt}
\begin{center}
\begin{tabular}{c c c c c }
\hline
 & Cable & Cable & Connector & Connector  \\
 &  Unroute & Route & Insertion & Insertion \\
 &  & &  & (Perception) \\
 \hline
CaP \citep{codeaspolicies2022} (Few-Shot) & 40\% & 0\% & 0\% & 0\% \\
\ours{}, FC (Few-Shot) & 80\% & 30\% & 20\% & 0\% \\
\ours{} (Zero-Shot) & 60\% &  0\% & 0\% & 0\% \\
\textbf{\ours{}} (Few-Shot) & \textbf{90\%} &  \textbf{100\%} & \textbf{90\%} & \textbf{60\%} \\
\hline
\end{tabular}
\end{center}
\caption{\small \textbf{RGMC~\citep{Sun2021ResearchCA} and NTB~\citep{Kimble2020BenchmarkingPF} Assembly Tasks.} On challenging manufacturing tasks, it's critical to have compliance and conditional terminations in the action space. \ours{} outperforms other LLM-based code-generation approaches for this reason.}
\label{tab:nist}
\end{table}

\textbf{Industrial Manipulation Tasks.}
To validate our method on more realistic force-based manipulation environments, we turn to the IROS RGMC~\citep{Sun2021ResearchCA} and NIST Assembly~\citep{Kimble2020BenchmarkingPF} experiments, which are arguably more directly targeted towards force-based manipulation than the peg insertion task. 
Indeed, we find it difficult to design an analogous scripted policy baseline for cable (un-)routing that would perform well across both tasks, which is why we omit it in this experiment.
From the results in Table~\ref{tab:nist}, we observe that \ours{} (Few-Shot) again consistently outperforms the baselines. Fixed Compliance (Few-Shot) is second-best, while Code-as-Policies \citep{codeaspolicies2022} (Few-Shot) performs worst, failing to complete the routing task even a single time. 
The Zero-Shot version of our method also fails for these tasks, and for two reasons: (1) the LLM tends to generate {\footnotesize\texttt{while}} loops that are incompatible with the way the API is structured and (2) the program is successful within a narrow range of force values that are difficult to infer without more information from examples. 

In the connector insertion task, we consider two cases: with and without fixed object placement. In the absence of fixed object placement, we estimate the plug and socket reference poses using pose estimators, which can reduce their accuracy by a small margin ($\epsilon \leq 4\mathrm{mm}$). Our method succeeded in maintaining a reasonable success rate in this case as well.

\label{sec:world_knowledge}
\begin{figure}
    \centering
    \includegraphics[width=0.5\textwidth]{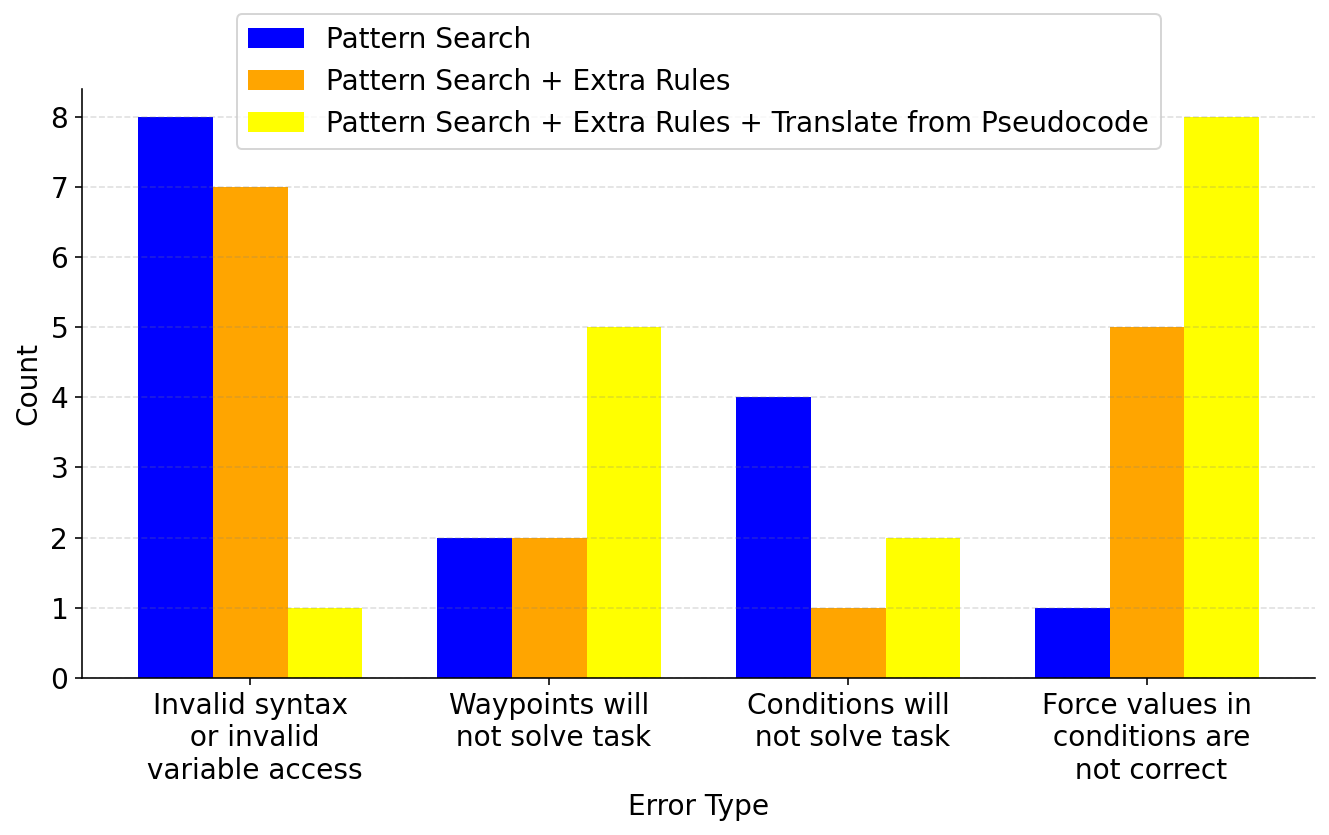}

  \caption{\small 
  We classify error types across three levels of hints. The severity of errors decreases from left to right. After combining all hints, the most common error is correct specification of force values.}
  \label{fig:error-classification}
\end{figure}

\subsection{Ablating prompt hints}
\label{sec:hint_ablation}
In the experiments underlying Sec.~\ref{sec:results}, we observed that incorporating additional hints is critical in eliciting relevant motion patterns.
To understand this phenomenon in depth, we plot the distribution over error types for increasing levels of hints on the cable un-routing task in the Zero-Shot setting.
We use three hint types: specifically asking for a pattern search, adding extra rules about accessing undefined variables, and asking the model to translate from pseudo-code, similar to Chain-of-Thought prompting~\citep{Wei2022ChainOT}.
Fig.~\ref{fig:error-classification} plots the distribution of failures for combinations of these hint types. We perform rejection sampling to estimate the likelihood of each error type given that the generated code fails. The errors are ordered from left to right based on the level of intervention required by a human operator to make the task succeed. For example, if the model outputs runnable code, but only moves the cable up without any wiggling motion, the operator will likely need to add new waypoints and tune the termination conditions and force values.
The most common example of invalid syntax or undefined variable accesses that we observe is in attempts to access the force values directly to construct {\footnotesize\texttt{for}} loops over waypoints. Explicitly adding rules against {\footnotesize\texttt{for}} loops helps somewhat, but asking the model to translate from pseudocode is the most helpful in minimizing these types of errors.

\section{Conclusions}

In this work, we study the capability of Large Language Models (LLMs) to generate policies for a variety of high-precision, contact-rich manipulation tasks. 
%
By allowing LLMs to place constraints on robot impedances and interaction forces, \ours{} improves success rates on subtasks derived from the Functional Manipulation Benchmark (FMB) and NIST Task Boards by 3x and 4x, respectively, when compared to code generation approaches that don't allow for compliance.
This is the first work to show that robotic code generation using language models can yield policies that are successful in completing contact-rich manipulation tasks.








\printbibliography
\break

\section*{APPENDIX}
We discuss implementation details in Sec.~\ref{sec:impelmentation_details}. We explore the fitness of language models for high-precision prediction in Sec.~\ref{sec:additional_experiments}. Robustness evaluations are in Sec.~\ref{sec:robustness}. Rollout visualizations are in Sec.~\ref{sec:rollout} and example prompts are in Sec.~\ref{sec:appendix-prompts}.

\section{Implementation Details}
\label{sec:impelmentation_details}

\subsection{Pose estimators}
We use a Faster-RCNN~\citep{renNIPS15fasterrcnn} model that does joint object detection and keypoint prediction with a ResNet-50 Feature Pyramid Network (FPN)~\citep{Lin2016FeaturePN} backbone. 
We train with a bounding box and keypoint regression loss mixture.

\subsection{Prompt Ablation Details}
To generate Fig.~\ref{fig:error-classification}, we take the first 15 prompts that generate an error (skipping any successful prompts) and classify them based on failure type with a human operator. Table~\ref{tab:nist} uses the best of five samples from the complete prompt.

\subsection{Compliant action space parameterizations}
It is worth noting that different robotic platforms offer different interfaces for control. Torque controlled robots, such as Franka-Emika Panda and Kuka LBR iiwa robots, readily lend themselves to the impedance control law described in Equation~\ref{eqn:impedance-control}.
On the other hand, position controlled robots, such as the UR5e, require a target position. This will affect how the compliant action space is parameterized.

To achieve soft interaction with the environment, we can implement an admittance controller that makes use of a force-torque sensor attached to our robot end-effector:
\begin{align*}
\ddot{x} = M^{-1}(K_p(x_d - x) + K_d(\dot{x}_d - \dot{x}) + (f_d - f_s))
\end{align*}
where $f_d$ is a desired force value, and $f_s$ is the current sensed force value. 
The use of admittance control enables us to parameterize our action space as $t_i = ([\mathbf{x}_{target}]_i, \sigma_i)$, where $[\mathbf{x}_{target}]_i$ denote target pose and $\sigma_i$ denote the stiffness vector.

Even though the admittance controller differs from the impedance controller described in Equation~\ref{eqn:impedance-control}, the parameters, namely $K_p$, $K_d$, and $M$, play a conceptually similar role to their corresponding variables in the impedance control law, which implies that our method can generalize across different robotic platforms.

\section{Probing language model fitness for precision tasks}
\label{sec:additional_experiments}

\subsection{Reasoning Over Continuous Spaces}
\label{sec:numerical_reasoning}
\begin{figure*}[h]
    \centering
    \begin{subfigure}{0.32\textwidth}
      \centering
      \includegraphics[width=\textwidth]{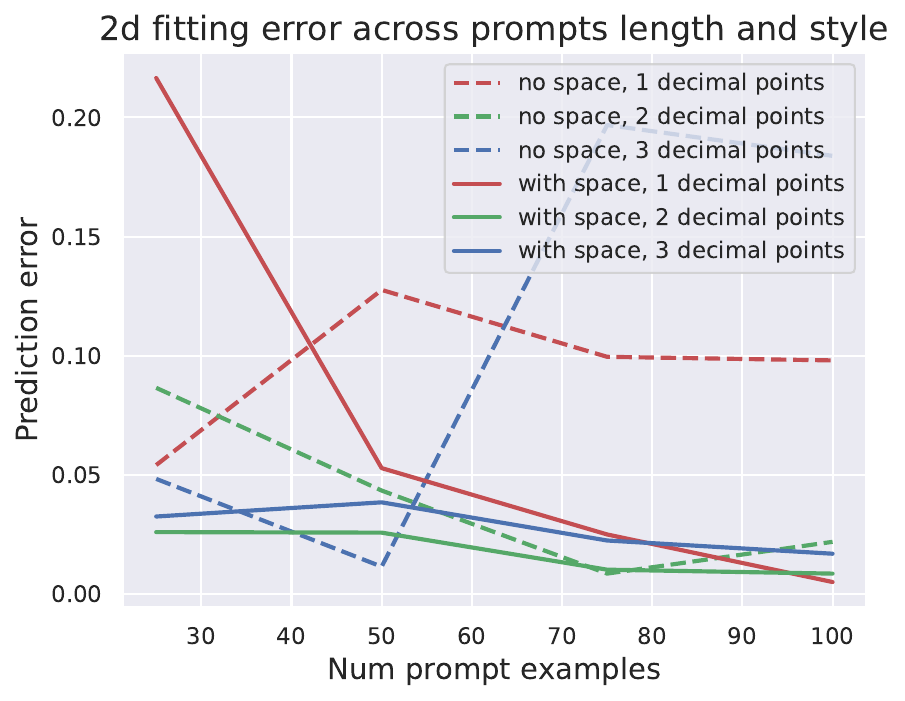}
      \caption{}
      \label{regresion}
      \end{subfigure}
     \begin{subfigure}{0.66\textwidth}
      \includegraphics[width=.5\columnwidth]{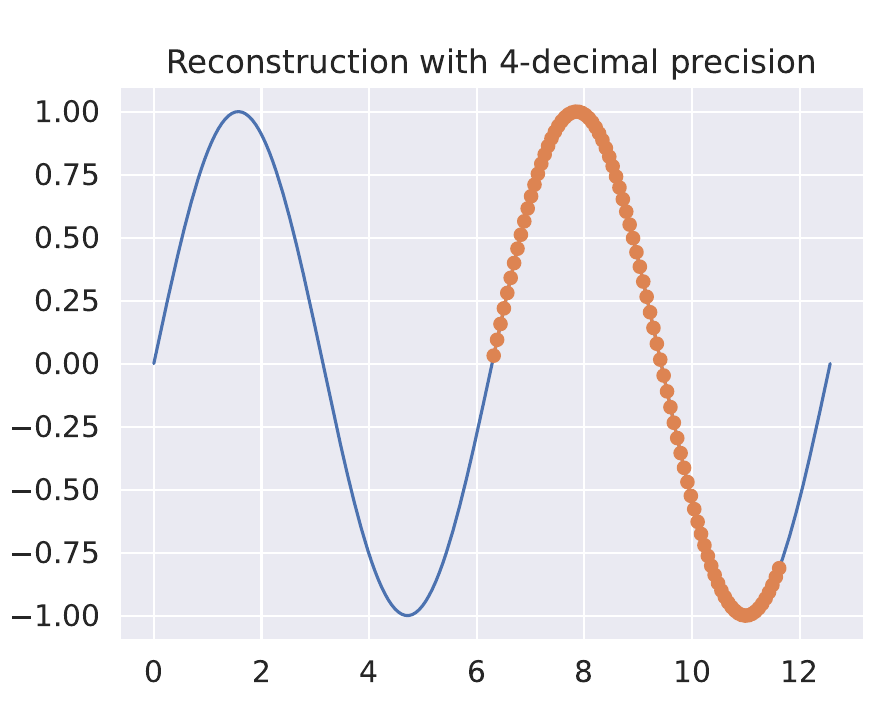}~
      \includegraphics[width=.5\columnwidth]{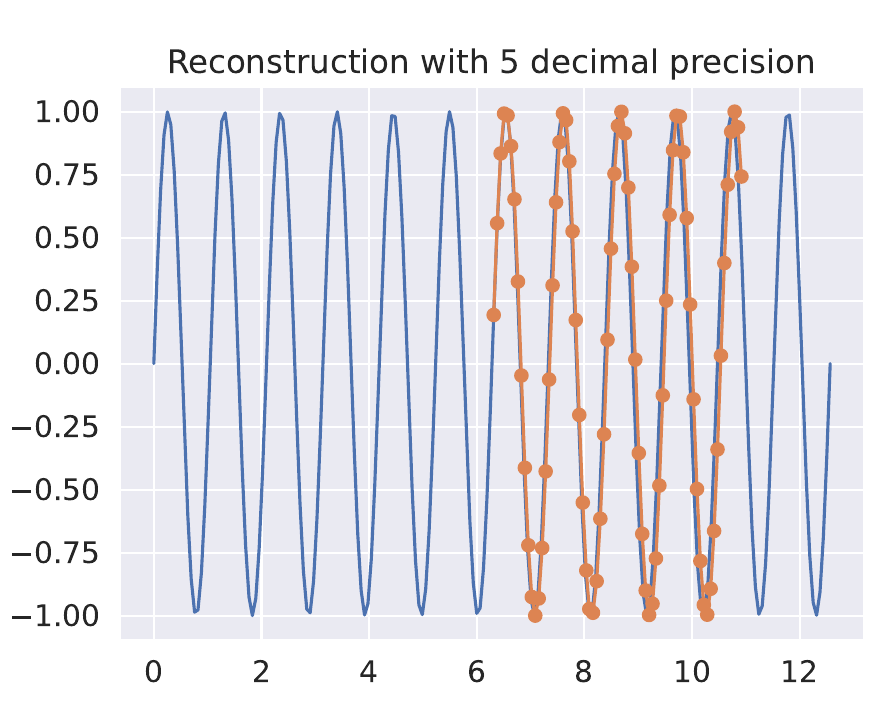}
      \caption{}
      \label{fig:reconstruction}
      \end{subfigure}

      \caption{\small (a) Reconstruction error across different floating point number resolutions and formatting styles. (b) Reconstructions of exemplary sinusoidal sequences for different floating point number resolutions.}
\end{figure*}
The goal of the main paper is to demonstrate that language models trained on vast, offline, internet datasets can still be useful in generalizing robotic policies. In Sec.~\ref{sec:results}, we show that this is because LLMs encode world knowledge about contact forces and information about object geometry that can be translated into motion patterns, such as the pattern search examples in Tab.~\ref{tab:fmb}. One reason this question is so pertinent is because of recent interest in amortizing the cost of robot policy learning by adapting language models to predict action tokens~\citep{brohan2023rt}. Code prediction is one way of using language models out of the box to yield generalizable robotic policies. However, for language models to be aqequate at action prediction for high-precision tasks, they also need to be able to extrapolate floating point sequences with high precision because contact-rich settings require millimeter-level precision.

While prior work demonstrates that LLMs are capable of serving as general pattern machines~\citep{mirchandani2023LargeLM}, they have not been shown to be able to reason over floating point numbers specifically. In this section, we thus examine two arithmetic tasks that we believe to be precursors to reasoning successfully in continuous action spaces: (1) a linear regression task, and (2) a continuous sequence extrapolation task. In both cases, and in contrast to prior work~\citep{mirchandani2023LargeLM}, we let the LLM operate on floating point numbers directly. \\

\subsubsection{Background}

\textbf{Arithmetic reasoning and pattern extrapolation abilities of LLMs}~\citet{mirchandani2023LargeLM} demonstrated that models trained on internet-scale data are capable of doing general spatial and sequential reasoning tasks. Similar to our work, their analysis is inspired by robot control-relevant reasoning problems. However, unlike them, we produce a proof of concept on real hardware and on a challenging contact rich manipulation task. Our arithmetic reasoning tasks are inspired by work from~\citet{garg2022what}, which shows that the Transformers~\citep{vaswani2017attentionIAYN} can learn simple function classes in-context after training on regression problems. Unlike them, we show that this capability emerges in models trained only on text generation. \\

\subsubsection{Results}
\textbf{Zero-Shot Regression In-Context:}
In Fig.~\ref{regresion}, we show the result of prompting a model with $x, y = f(x)$ from a 2-dimensional linear function $f$. We quantify the ability of the model to implicitly learn the linear function by measuring the $\hat{y}$ output by the model for a given $x$, and plotting the resulting error for different floating point number resolutions.
The x-axis denotes the number of x, y pairs provided to the LLM and the y-axis is the average prediction error of $\hat{y}$ for a given value $x$.
We compare two different formatting strategies. In the first, the $x,y$ pairs are written out as {\footnotesize\texttt{f(x)=y}}. The second formatting strategy adds spaces between each digit so that each digit is treated as a separate token. For example, {\footnotesize\texttt{f(1.393)=4.107}} is formatted as {\footnotesize\texttt{f(1.3 9 3)=4.1 0 7}}. As can be seen in Fig.~\ref{regresion}, space-formating with GPT-4 is able to successfully regress to the targets within an error of $0.01$ when using at least $3$ decimal places.

\textbf{Sequence Extrapolation:} In Fig.~\ref{fig:reconstruction}, we illustrate the ability of an LLM to extrapolate sequences of two-dimensional series of numbers (concretely, sinusoids of different frequencies), as this might transfer to path-following behaviors required in the robotic context. 
To generate these plots, we feed samples of the sinusoidal function up to a certain value and then auto-regressively sample from the LLM. The sampled points are shown in orange. For visualization purposes, the true function is shown in blue.
Qualitatively, we see in the figure that the LLM is indeed capable of extrapolating the chosen sinusoids. Unlike~\citet{mirchandani2023LargeLM}, we find that this ability also works for floating point numbers. To realize these results, we modify the tokenization strategy by placing a space in between each digit. 
The extrapolation ability begins to break down as the precision and frequency of the sinusoid increase, which can be seen on the right-most side of Fig.~\ref{fig:reconstruction}.


The takeaway from this set of experiments is that LLMs trained entirely on offline language data can indeed achieve reasonable performance on arithmetic tasks on continuous spaces.
This result shows that large language models are capable of performing least squares up to a small degree of error with in-context learning.

\subsection{Reasoning over spatial patterns}
To protoype alternative ways of providing the language model with perceptual information for contact rich tasks, we provide schematics of an assembly board in natural language. In Sec.~\ref{sec:tasks}, we allow the language model to reason over object poses via calls to {\footnotesize\texttt{estimate\_and\_update\_pose}}. Alternatively, we could pass tokenized versions of the scene into the language model~\cite{gal2022textual,liu2023llava}. We explore this by prompting with spatial patterns.
\begin{minipage}[t]{\linewidth} 
\begin{lstlisting}[title={Spatial Patterns}, language=Python, linebackgroundcolor={%
    \ifnum\value{lstnumber}>100
        \color{codehighlight!25}
    \fi
    }
    ]
"""
Here is the board layout:
    B
ccc c
cscsc
c ccc
1

x translations move left to right and y translations 
moves bottom to top. The distance between screws is 
.03M.
"""
\end{lstlisting}
\label{fig:spatial-patterns}
\end{minipage}%

\textbf{Spatial patterns} are symbolic summaries of a given scene and act as a character-based representation of what is visible to the agent. An example of a symbolic representation of an IROS RGMCS~\citep{Sun2021ResearchCA} board is summarized in Figure~\ref{fig:spatial-patterns}. The \texttt{c} symbol refers to the path of the cable, \texttt{S} refers to a screw, and \texttt{B} refers to a plastic channel component through which the cable is routed. These would also be defined in the prompt. Although these are specified by the prompt designer in our examples, they could also be generated by a vision-language model or other perception APIs~\citep{wi2023calamari}.

\begin{figure}
    \centering
    \includegraphics[width=0.15\textwidth]{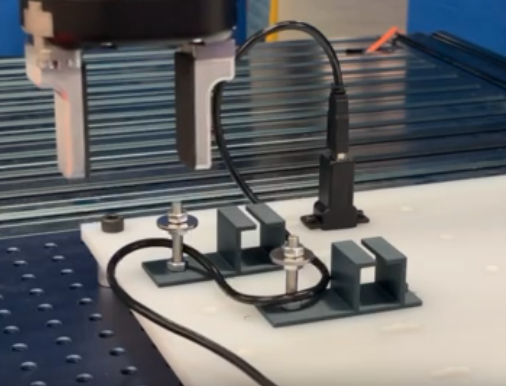}
    \includegraphics[width=0.15\textwidth]{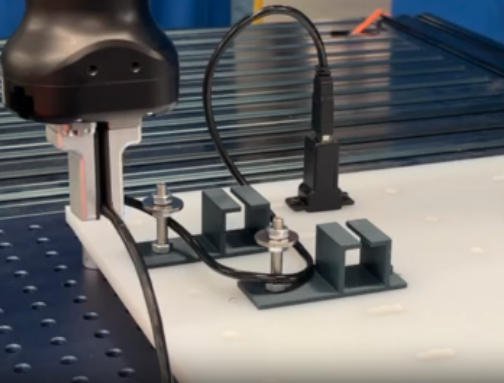}
    \includegraphics[width=0.15\textwidth]{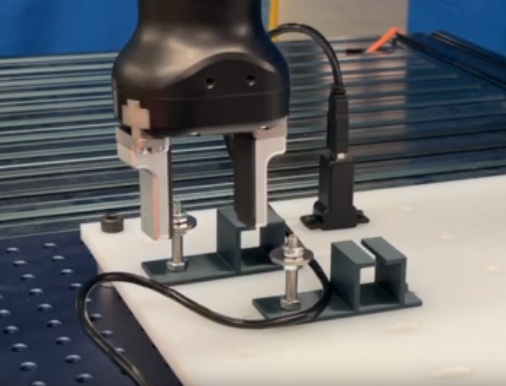}

  \caption{\small 
  We prototype an alternative way of passing perceptual information into the language model on the cable unraveling task from IROS RGMC~\cite{Sun2021ResearchCA}. This is a typical rollout.}
  \label{fig:unravel-task}
\end{figure}

In addition to adding the spatial pattern to the prompt, we append the following task command: {\footnotesize\texttt{Explain how to unravel the cable from the screws and then generate the code to unravel the cable if the robot is currently holding the cable at position 1.}}Using \ours{}, the best code generation is successful in removing the cable from the first screw in 8 out of 10 evaluation runs but only successful in removing the cable from both screws in 2 out of 10 cases. We expect that these numbers could be further improved by allowing for regrasps or by implementing more advanced and iterative scene tokenization strategies, which we believe is a strong area for future work.

\begin{figure*}[h]
    \centering
    \includegraphics[width=\textwidth]{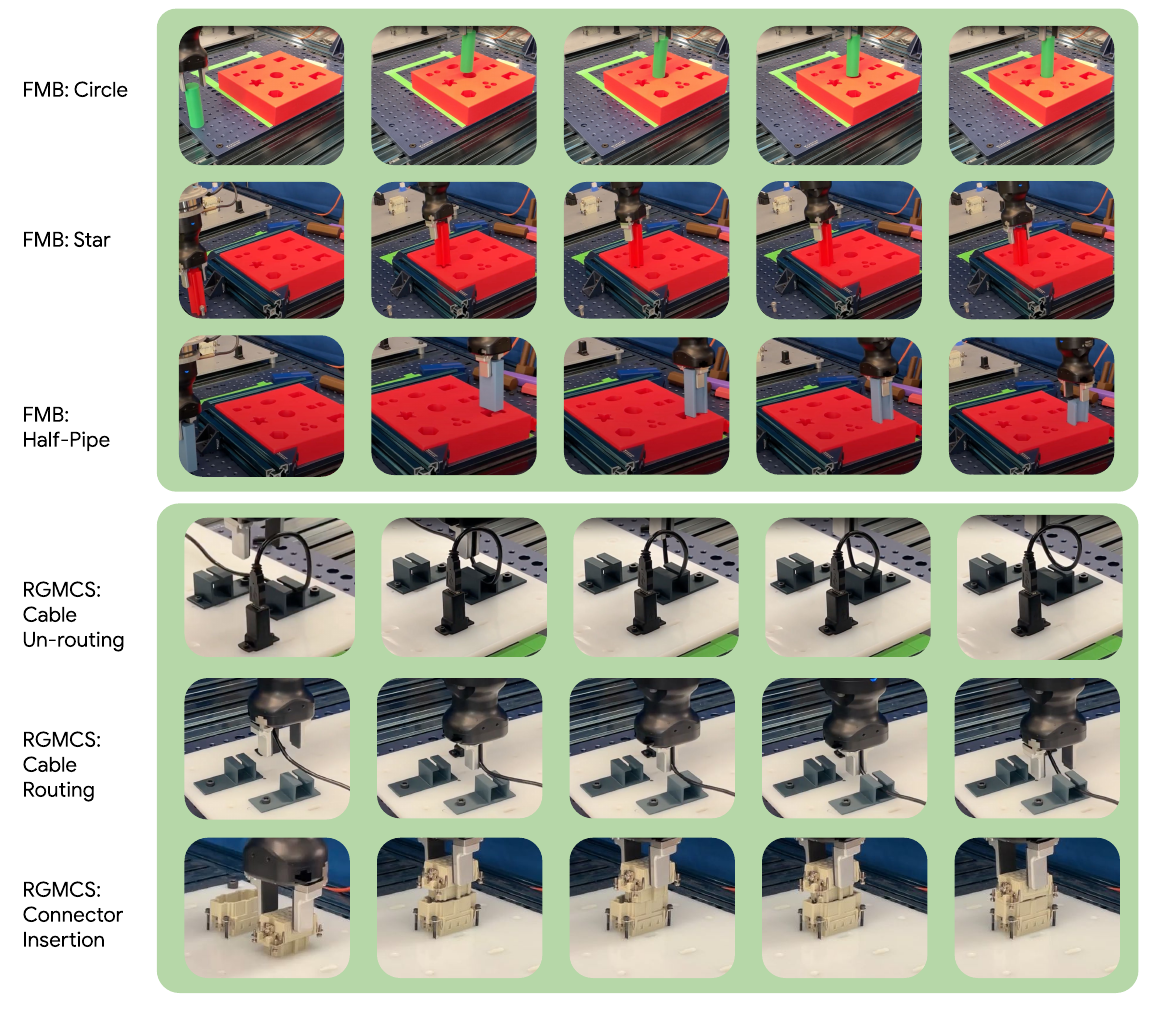}
    \caption{\small Example rollouts on all tasks.}
    \label{fig:rollouts}
\end{figure*}

\section{Robustness}
\label{sec:robustness}
\subsection{Cable un-routing}
We study the robustness to initial position on the cable unrouting task. The starting position on this task is parameterized by a script that localizes and picks up the cable, which is assumed to be threaded through one of the RGMC plastic components. To make the task challenging, the gripper starting position is offset in the positive or negative x direction, which corresponds to the left and right side of the tunnel. In Tab.~\ref{tab:unrouting-robustness} we show that \ours{} is able to successfully un-route the cable regardless of the displacement direction.
\begin{table}[t]
\label{tab:your-label}
\begin{center}
\begin{tabular}{l c c}
\hline
 & Left-side tunnel & Right-side tunnel \\ 
\hline
CaP~\cite{codeaspolicies2022} (Few-Shot) & 60\% & 20\% \\
\ours{}, FC (Few-Shot) & 60\% & \textbf{100\%} \\
\textbf{\ours{}} (Few-Shot) & \textbf{100\%} & 80\% \\
\hline
\end{tabular}
\end{center}
\caption{\small Cable unrouting~\citep{Sun2021ResearchCA} from different starting positions.}
\label{tab:unrouting-robustness}
\end{table}

\section{Rollout visualization}
\label{sec:rollout}
We visualize rollouts from our method in Fig.~\ref{fig:rollouts}.

\section{Prompt and Output Code Example}
\label{sec:appendix-prompts}
In this section, we provide a complete prompt example for the cable un-routing task (Listing~\ref{lst:code-example-1}) and the GPT-4-generated output (Listing~\ref{lst:code-example-2}). Importantly, this example demonstrates how that the language model is capable of generating novel combinations of moves and constraint conditions. In the prompt, there is no constraint on right-ward snags for poses moving to the left in the prompt, but the language model is able to synthesize these constraints for the task. Please see our paper website for the remaining prompts.

\begin{lstlisting}[float=*, float=*, caption={Complete prompt example for the cable un-routing task},captionpos=t, label={lst:code-example-1}]
    prompt = """
You're a robot holding a cable that's threaded through a tunnel with a small opening at the top.
You need to unroute the cable by removing it from the tunnel.

You will have access to the following methods, which are imported directly:
- cartesian_admittance_move: This moves the robot to a target_pose until a termination condition is reached.
  Args:
      max_cartesian_stiffness:
          The maximum allowed stiffness along each cartesian dof (6d), expressed in
          the robot base frame.
      target_impedance:
          (0,1] 6d-vector specifying the target impedance along each cartesian dof.
      target_pose:
          Target pose for the robot flange frame in the base frame.
      termination_condition:
          Termination condition.
      virtual_cartesian_inertia:
          The diagonal representation of the desired virtual Cartesian inertia
          matrix, expressed in the robot base frame [kg, kg m^2]
      execution_timeout_seconds:
          Timeout for execution. Defaults to 30s if not specified
          Default value: 10.0
      tare_ft_sensor: False when in contact, True otherwise.
- types_pb2.Comparison: this specifies the termination condition above. It can't be accessed directly, it can only be passed as an argument to the cartesian_admittance_move method.
  Args:
      operation: types_pb2.Comparison type. One of GREATER_THAN_OR_EQUAL, LESS_THAN_OR_EQUAL, APPROX_EQUAL, or APPROX_NOT_EQUAL.
      state_variable_name: Variable where condition is applied. One of policy.status.{x_force, y_force, z_force, translation_error, rotation_error}.
      double_value: [-0.4, 0.4] value of state variable on which operation is applied
- types_pb2.Condition: Can't be accessed directly, can only be passed as an argument to the cartesian_admittance_move method.
  Args:
      comparison: types_pb2.Comparison type.
- types_pb2.ConjunctionCondition: Wrapper to compose multiple conditions together.
  Args:
      operation: types_pb2.ConjunctionCondition type. One of ALL_OF or ANY_OF .
      conditions: A list of Conditions over which the operation is applied
- pose3_rpy: returns a pose object that specifies a pose. Note: Poses can be composed using the multiply method. E.g., pose1.multiply(pose2) applies pose2 on top of pose1.
  Args:
      translation: x, y, z translation.
      rotation: rotation along x, y, and z.
- move_gripper: opens and closes the gripper.
  Args:
      percent_open: Controls how open gripper is. 0 is fully closed, 100 is fully open.
- grasp_cable: grab the free edge of the routed cable

You will also have access to the following fields:
- pose(number): the robot pose just above number. You can build more poses by using this as a reference with the multiply method.
"""
\end{lstlisting}

\begin{lstlisting}[float=*]
"""
  Args:
      number: the pose number for the layout below.
- grasp(pose): go to grasp pose and grasp the cable.
  Args:
      pose: the robot pose above the grasp position

Rules:
- Don't define any new methods
- Don't call any undefined methods
- Don't add any if statements or while loops

Here are some examples using the cartesian_admittance_move method while holding the cable:

```python
position_condition = types_pb2.Comparison(
  operation=types_pb2.Comparison.LESS_THAN_OR_EQUAL,
  state_variable_name="policy.status.translation_error",
  double_value=0.001,
)

# move down until contact is reached
upward_force_threshold = 0.4
contact_condition = types_pb2.Condition(
    comparison=types_pb2.Comparison(
        operation=types_pb2.Comparison.GREATER_THAN_OR_EQUAL,
        state_variable_name="policy.status.z_force",
        double_value=upward_force_threshold,
    )
)
down_pose = pose(1).multiply(pose3_rpy(translation=[0, 0, -.01]))
cartesian_admittance_move(
  target_pose=down_pose,
  termination_condition=termination_condition=types_pb2.ConjunctionCondition(
        operation=types_pb2.ConjunctionCondition.ANY_OF,
        conditions=[contact_condition, position_condition]
    ))

# move up unless snag is detected
downward_force_threshold = -0.4
snag_condition = types_pb2.Condition(
    comparison=types_pb2.Comparison(
        operation=types_pb2.Comparison.LESS_THAN_OR_EQUAL,
        state_variable_name="policy.status.z_force",
        double_value=downward_force_threshold,
    )
)
up_pose = pose(1).multiply(pose3_rpy(translation=[0, 0, .01]))
cartesian_admittance_move(
  target_pose=up_pose,
  termination_condition=termination_condition=types_pb2.ConjunctionCondition(
        operation=types_pb2.ConjunctionCondition.ANY_OF,
        conditions=[snag_condition, position_condition]
    ))

# move the cable to the right unless snag is detected
leftward_force_threshold = 0.4
leftwardsnag_condition = types_pb2.Condition(
    comparison=types_pb2.Comparison(
        operation=types_pb2.Comparison.GREATER_THAN_OR_EQUAL,
        state_variable_name="policy.status.x_force",
        double_value=leftward_force_threshold,
    )
)
right_pose = pose(1).multiply(pose3_rpy(translation=[-0.01, 0, 0]))
cartesian_admittance_move(
  target_pose=up_pose,
  termination_condition=termination_condition=types_pb2.ConjunctionCondition(
        operation=types_pb2.ConjunctionCondition.ANY_OF,
        conditions=[leftwardsnag_condition, position_condition],
    ))
```

Currently the robot is holding the cable at pose(1), but the cable is stuck in a tunnel with a small opening at the top. Describe a strategy for unrouting the cable using force information to search for the opening and then write a program using the APIs above to unroute the cable and lift it out of the tunnel."""

messages = [{"role": "user", "content":prompt}]
response = openai.ChatCompletion.create(
    model="gpt-4",
    messages = messages,
    temperature=0.0,
    max_tokens=512*3,
    frequency_penalty=0.0
)
print(response["choices"][0]["message"]["content"])
\end{lstlisting}

\begin{lstlisting}[float=*, caption={GPT-4-generated output for the cable-unrouting task},captionpos=b, label={lst:code-example-2}]
    Strategy:
1. First, we will move the cable upwards until we detect a snag. This will help us to find the top of the tunnel.
2. Once we detect a snag, we will move the cable to the right until we detect another snag. This will help us to find the right wall of the tunnel.
3. After finding the right wall, we will move the cable to the left until we no longer detect a snag. This will help us to find the opening of the tunnel.
4. Finally, we will move the cable upwards again to lift it out of the tunnel.

Program:

```python
# Step 1: Move up until snag is detected
downward_force_threshold = -0.4
snag_condition = types_pb2.Condition(
    comparison=types_pb2.Comparison(
        operation=types_pb2.Comparison.LESS_THAN_OR_EQUAL,
        state_variable_name="policy.status.z_force",
        double_value=downward_force_threshold,
    )
)
up_pose = pose(1).multiply(pose3_rpy(translation=[0, 0, .01]))
cartesian_admittance_move(
  target_pose=up_pose,
  termination_condition=types_pb2.ConjunctionCondition(
        operation=types_pb2.ConjunctionCondition.ANY_OF,
        conditions=[snag_condition, position_condition]
    ))

# Step 2: Move right until snag is detected
leftward_force_threshold = 0.4
leftwardsnag_condition = types_pb2.Condition(
    comparison=types_pb2.Comparison(
        operation=types_pb2.Comparison.GREATER_THAN_OR_EQUAL,
        state_variable_name="policy.status.x_force",
        double_value=leftward_force_threshold,
    )
)
right_pose = pose(1).multiply(pose3_rpy(translation=[-0.01, 0, 0]))
cartesian_admittance_move(
  target_pose=right_pose,
  termination_condition=types_pb2.ConjunctionCondition(
        operation=types_pb2.ConjunctionCondition.ANY_OF,
        conditions=[leftwardsnag_condition, position_condition],
    ))

# Step 3: Move left until no snag is detected
rightward_force_threshold = -0.4
rightwardsnag_condition = types_pb2.Condition(
    comparison=types_pb2.Comparison(
        operation=types_pb2.Comparison.LESS_THAN_OR_EQUAL,
        state_variable_name="policy.status.x_force",
        double_value=rightward_force_threshold,
    )
)
left_pose = pose(1).multiply(pose3_rpy(translation=[0.01, 0, 0]))
cartesian_admittance_move(
  target_pose=left_pose,
  termination_condition=types_pb2.ConjunctionCondition(
        operation=types_pb2.ConjunctionCondition.ANY_OF,
        conditions=[rightwardsnag_condition, position_condition],
    ))

# Step 4: Move up to lift the cable out of the tunnel
cartesian_admittance_move(
  target_pose=up_pose,
  termination_condition=types_pb2.ConjunctionCondition(
        operation=types_pb2.ConjunctionCondition.ANY_OF,
        conditions=[snag_condition, position_condition]
    ))
```
\end{lstlisting}

\end{document}